# Extended Multi-stream Temporal-attention Module for Skeleton-based Human Action Recognition (HAR)

Faisal Mehmood[1], Xin Guo[1], Enqing Chen[1,2*], Muhammad Azeem Akbar[3], Arif Ali Khan[4] and Sami Ullah[5]

[1]School of Electrical and Information Engineering, Zhengzhou University, Zhengzhou, 450001, Henan, China.
[2]Henan Xintong Intelligent IOT Co., Ltd., No. 1-303 Intersection of Ruyun Road and Meihe Road, Zhengzhou, 450007, Henan, China.
[3]Department of Software Engineering, LUT University, Lahti, 15100, Finland.
[4]M3S Empirical Software Engineering Research University of Oulu, Oulu, 90570, Finland.
[5]Department of Computer Science, Government College University Faisalabad, Faisalabad 38000, Punjab, Pakistan.

*Corresponding author(s). E-mail(s): ieeqchen@zzu.edu.cn;
Contributing authors: faisalmehmood685@uaf.edu.pk; iexguo@zzu.edu.cn; azeem.akbar@lut.fi; arif.khan@oulu.fi; samiullah@gcuf.edu.pk

## Abstract

Graph convolutional networks (GCNs) are an effective skeleton-based human action recognition (HAR) technique. GCNs enable the specification of CNNs to a non-Euclidean frame that is more flexible. The previous GCN-based models still have a lot of issues: (I) The graph structure is the same for all model layers and input data. GCN model's hierarchical structure and human action recognition input diversity make this a problematic approach; (II) Bone length and orientation are understudied due to their significance and variance in HAR. For this purpose, we introduce an Extended Multi-stream Temporal-attention Adaptive GCN (EMS-TAGCN). By training the network topology of the proposed model either consistently or independently according to the input data, this data-based technique makes graphs more flexible and faster to adapt to a new dataset. A spatial, temporal, and channel attention module helps the adaptive graph convolutional layer focus on joints, frames, and features. Hence, a multi-stream framework representing bones, joints, and their motion enhances recognition accuracy. Our proposed model outperforms the NTU RGBD for CS and CV by 0.6% and 1.4%, respectively, while Kinetics-skeleton Top-1 and Top-5 are 1.4% improved, UCF-101 has improved 2.34% accuracy and HMDB-51 dataset has significantly improved 1.8% accuracy. According to the results, our model has performed better than the other models. Our model consistently outperformed other models, and the results were statistically significant that demonstrating the superiority of our model for the task of HAR and its ability to provide the most reliable and accurate results.

## 1. Introduction

Skeleton-based Human Action Recognition (HAR) plays an essential role in many applications, including video surveillance, robotics, and human-computer interaction (HCI) [1, 2]. Graphs of human skeleton data illustrate human movements more clearly than RGB video data. Skeletal data is graphed, so viewpoint variations, occlusions, background clutter, inter-class position discrepancies, lighting, and clothing do not affect it. Human skeleton data can be collected via Kinect, motion sensors, or pose prediction algorithms [3, 4]. Skeleton-based HAR is becoming more popular than older methods that use RGB data for detection. This is because it can quickly adapt to complicated situations and backgrounds. At the beginning of deep learning for skeleton-based HAR, skeleton data had to be carefully organized into joint-coordinate vectors. These methods take a long to train and have implications because the models are very complicated [5]. The non-Euclidean structure of the skeletal data, which displays the spatial connections between the body's joints, is also not adequately exploited by these methods.

Graph Convolutional Networks (GCNs) have successfully succeeded in several HAR tasks [6, 7], because they adapt the convolution process from grid data to graph data structures. Recently, significant progress has been made in using GCNs for skeleton-based HAR [8, 9]. These techniques employ skeletons to represent a sequence of in-body positions, each representing an action. Skeleton data is processed as a graph to show the spatial relationship between body joints, and skeleton sequences show action time dynamics. Most current GCN-based techniques use deep feed-forward networks to simulate the skeleton's temporal characteristics. However, when constructing the input graph, most GCNs only consider the skeleton's natural joint connections. Consequently, they miss essential joint information inferred from connections between distant joints that are not linked.

This paper addresses these challenges by developing an extended multi-stream temporal-attention adaptive GCN (EMS-TAGCN). This research presents two types of adaptable graphs used in GCN. The first graph is the general graph. The dataset information is used to construct the graph to determine the adjacency matrix to create the graph. As a result, this graph structure is more effective in recognizing human activity as a body-based graph. This second one is a single graph whose sides are made up of similar points. The large number of data samples allows the module to store a unique structure for each input. By utilizing a gating mechanism, it is possible to merge the two different types of graphs, changing the significance of each in every model layer. It's vital to remember that each graph is optimized separately across layers; this method may perform better with neural networks' hierarchical nature.

Hence, this data-driven strategy makes the graph-making mechanism more adaptive to varied datasets. In addition, numerous tasks require the attention strategy, which is currently effective [10, 11]. From a spatial perspective, a specific action is frequently associated with and defined by a critical subset of the joints. From a temporal perspective, there may be more than one stage in an action flow, with various substages or frames essential to the outcome. From the viewpoint of the features, a convolutional feature map comprises several channels, each with a different level of importance. Each stream performs a specific task to the actions and data samples it handles. Our main objective is to build an STC-attention module that can adapt activations of joints, frames, and channels according to experimental data samples. We get inspiration from these results. The module can be added to the convolutional layer of any graph. It only has a few options, but it can still help make things run better.

We run extensive experiments on four massive datasets to prove that the developed (EMS-TAGCN) model outperforms others. These datasets are NTU-RGBD [12], Kinetics Skeleton [13], UCF-101 [14] and HMDB-51 [15]. Regarding skeleton-based HAR, our model performs best across both datasets. We also present the model's learning attention maps and adaptive graphs to show how the four modalities work. While skeletons provide more information about an object's look than RGB data, we compare the two and recommend merging them via skeleton-guided cropping.

## 1.1 Research Objectives

The primary objective of our research is to develop an advanced graph convolutional network (GCN) model that enhances the accuracy and adaptability of skeleton-based action recognition systems. This goal will be achieved through the following specific aims:

- To integrate multiple modalities of skeletal data, including joint positions, bone lengths, orientations, and motion details, into the proposed GCN architecture to enrich feature representation.
- To investigate and implement mechanisms to dynamically adapt the graph topology of the GCN, allowing for more flexible and context-aware action recognition.
- To explore the incorporation of spatial-temporal-channel attention modules within the GCN framework to better capture informative features and improve discriminative power.
- To conduct comprehensive evaluations of the proposed Extended Multi-stream Temporal-attention Adaptive GCN (EMS-TAGCN) model on diverse datasets, such as NTU-RGBD Kinetics-skeleton, UCF-101 and HMDB-51 to validate its effectiveness and generalizability across different scenarios and domains.

Through these objectives, this research aims to push the boundaries of what is currently achievable with GCN-based models in action recognition, addressing specific gaps in the literature while setting new standards for performance and adaptability in the field.

### 1.2 Research Contributions

This research proposes a new graph convolutional network (GCN) architecture to recognize human action from skeleton data. Based on skeleton data, the model is expected to produce competitive results compared to the current state-of-the-art methods. Our contributions lie in four folders:

- The proposed EMS-TAGCN model more accurately represents HAR by employing joint-to-body-part components. It is the first to model temporal correlations among graph features using the graph convolutional module to generate a highly discriminate pose-related representation HAR identification.
- We create a gating mechanism to dynamically change the two graphs' relative importance. Our second contribution is an STC-attention module that has shown to be a powerful resource for facilitating the model's courtesy toward important joints, frames, and features.
- The model was expanded to a multi-stream architecture with diverse bone and joint movements. We do more extensive trials and explain them to demonstrate the usefulness and necessity of the intended modules and data formats.
- The proposed model outperforms on both datasets: NTU-RGBD improved by CS 0.4% and CV 1.4%, while on Kinetics-skeleton, Top-1 and Top-5 are 1.4% improved, UCF-101 has improved 2.34% accuracy and HMDB-51 dataset has significantly improved 1.8% accuracy. Finally, we offer a robust pose-guided cropping method that works well in RGB. The goal is to integrate skeletons and modalities.

The remaining papers are organized as follows: Section 2 discusses the literature review. Section 3 presents the mechanisms of the introduced model. The experimental setting for our work is discussed in Section 4. Section 5 presents the analysis results, RGB modalities, and comparison with the latest methods. The conclusion of this research is presented in final section 6.

## 2. Review of Literature
### 2.1 Skeleton-based HAR

In the past few years, skeletons have been used in many studies on action recognition. We only talk about important works here. In skeleton-based HAR, handcrafted features are often used to describe the human body [16, 17]. The translations and iterations of the skeleton's joints can be encoded by using [18, 19]. B. Fernando et al. [20] customize the place-gathering method to define the evidence using the top-rank attributes.

Deep learning models that rely on data, such as RNNs and CNNs, are now the most commonly used methods. Conventional wisdom holds that skeletal data is best understood as a series of spatial and temporal vectors representing a human body joint [21]. Y. Du *et al.* [22] introduce hierarchical bidirectional recurrent neural network (RNN) models that detect skeleton sequences that partition the human body into discrete components before transmitting them to multiple sub-networks. X. Wang and H. Deng [23] incorporates a component for temporal consideration into the LSTM-based model so that the system can autonomously monitor the skeleton series' temporal classifier region. P. Zhang *et al.* [21] combine a view conversion method with an LSTM-based system that changes skeleton data for HAR. C. Si *et al.* [24] proposed a methodology that utilized timed stack learning (TSL) and reasoning about space (SRN). This approach involves the SRN gathering data on the interrelationships among different body parts while the TSL characterizes the exact temporal dynamics.

## 2.2 Graph Convolutional Networks (GCN)

GCN uses convolutional methods on graph data instead of image data like traditional CNNs [25]. Because they get good results on graph data, many experts have used GCN in various situations. GCN can be broken down into two groups: spectral and spatial. The spectral GCN changes the graph in the spectral domain and uses the Fourier transform on the graph. The spatial GCN, on the other hand, gets information from nodes that are close by. Spectral GCN is used in the method suggested in this work.

Traditional convolutional neural networks (CNNs) commonly take flat, regular grids as input and process audio, video, and visual inputs. Graph data, on the other hand, is notoriously difficult to describe using CNN due to its inherent size and shape variability. Parse trees, social networks, and molecular networks are more common in the actual world. Over the past ten years, a lot of research has been done on the application of GCNs to perform operations on graphs. When compared to classic convolutional neural networks (CNNs), GCNs are superior since they can combine images into graphs of any shape or size [1, 26]. When building GCNs, two main lines of reasoning are considered: spatial and spectral. The graph vertices and their associates are effectively convolution-zed using spatial perspective methods. Due to the lack of intrinsic structure in a graph's nodes and edges, it isn't easy to construct communities from them. This method finds nearby mates using criteria manually created by [6, 27, 28]. M. Niepert *et al.*, [27] Choose neighbors of the vertices based on the distances in the graph. A method for reducing unnecessary fake vertices and superfluous pads is presented. X. Wang and A. Gupta *et al.* [29] construct a video graph containing all the observable objects and individuals for action recognition. As a result of feature similarity and spatial-temporal correlations, the immediate surroundings of each vertex are defined.

When spectral viewpoint techniques are used instead of spatial perspective techniques, they use graph Laplace matrices' eigenvalues and eigenvectors. The Fourier transform is a mathematical concept that merges different graphs in the frequency domain to represent a signal in terms of its frequency components. [26]. It does this by not needing to remove locally linked areas from graphs at each compression step [1, 26]. D. I. Shuman *et al.*, [26] compared to previous polynomial filters, recurrent Chebyshev polynomials provide a significantly more realistic filtering method [1, 30]. This method was further improved by using the first-order estimate of the spectral graph convolutions. This work makes use of spatial perspective techniques.

## 2.3 Spatio-Temporal GCN (ST-GCN)

GCN has been the subject of substantial research into skeleton-based human activity recognition. The most recent and notable research has focused on spatio-temporal GCN to detect changes in time and the relationships between different events in a video. The standard ST-GCN architecture uses skeleton graphs and a series of ST-GCN blocks that perform spatial and temporal graph convolutions in alternating fashion [25]. The action class is ultimately predicted by fully connected dense layers and the SoftMax classifier.

W. Zheng *et al.* [31] introduce a novel framework called ST-GCN. The OpenPose human pose estimation technique extracts skeleton data from videos and converts it to skeleton graphs. ST-GCN extracts temporal and spatial data for action categorization using a fixed static graph and a tri-categorical partitioning approach. Yang *et al.* [32] create a dynamic skeleton graph and three-set partitioning using a data-driven learning approach. For every skeleton frame, the attention-based adjacency matrix is used to apply spatial features. At the same time, the velocity semantic information is used to extract the temporal aspects.

# 3. EMS-TAGCN Model

This subsection comprehensively explores the EMS-TAGCN components (Extended Multi-stream Temporal-attention Adaptive GCN). A detailed examination is conducted on the multi-stream architecture, AGCN, and the Attention module. The latter includes an in-depth discussion on spatial, temporal, and Channel attention modules within this section.

## 3.1 Extended Multi-stream Network

Many current skeleton-based graph convolution methods only use information about joints and bones and don't include how they move, how much distance between joints, and the length between bones. The graph convolution network can get higher-order information from multiple convolutions. However, using the higher-order information directly as input can strengthen the recognition effect. So, this study looks at the multi-stream data, the motion data, and the length and distance data from bones and joints at the same time. The space between two adjacent joints within the same frame constitutes a bone. Proximal point $v_i = (x_i, y_i, z_i)$ denotes the adjacent joints closer to the skeleton's center of gravity. The distal point $v_j = (x_j, y_j, z_j)$ denotes the adjacent joints at a greater distance. Every bone is shown as a vector $e_{i,j} = (x_i \rightarrow x_j, y_i \rightarrow y_j, z_i \rightarrow z_j)$, with $x_i$ being the closest and $y_i$ being the farthest. The skeleton has no closed loop, so there is one less than the joints. For data integrity, an empty bone is added so that the data from the joints and the bones can use the same graph structure and network. The difference between two frames of the same joint or bone is known as a joint or bone movement. In the *f* frame, a shared data point is $v_{i,f} = (x_{i,f}, y_{i,f}, z_{i,f})$. The joint motion data is represented as follows: $v_{mi,f\_f+1} = (x_{i,t+1} - x_{i,f}, y_{i,f+1} - y_{i,f}, z_{i,f+1} - z_{i,f})$ in the following frame, $v_{i,f+1} = (x_{i,f+1}, y_{i,f+1}, z_{i,f+1})$. In the present frame, the bone data consists of $e_{ij,f} = (\hat{x}_{ij,f}, \hat{y}_{ij,f}, \hat{z}_{ij,f})$. It is $e_{ij,f+1} = (\hat{x}_{ij,f+1}, \hat{y}_{ij,f+1}, \hat{z}_{ij,f+1})$ in the following frame. $e_{ij,f\_f+1} = (\hat{x}_{ij,f+1} - \hat{x}_{ij,f}, \hat{y}_{ij,f+1} - \hat{y}_{ij,f}, \hat{z}_{ij,f+1} - \hat{z}_{ij,f})$ are possible ways to display the bone motion data. Our proposed model architecture is shown in Figure 1.

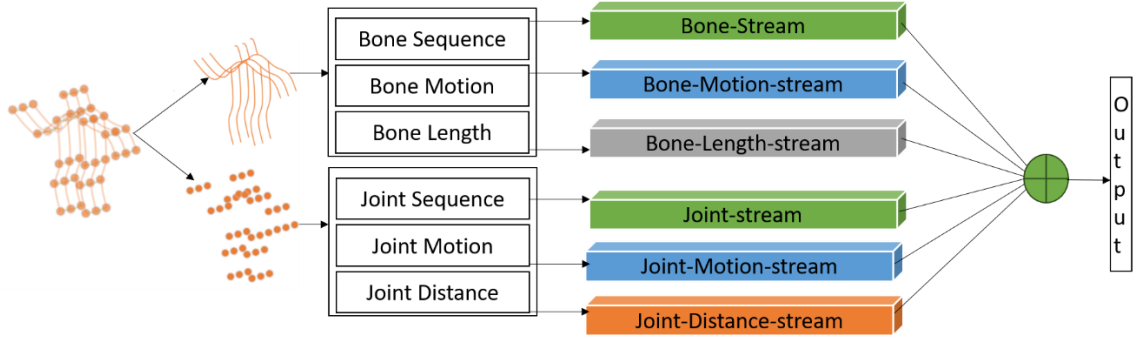

**Figure 1:** EMS-TAGCN network architecture.

### 3.2 Adaptive Graph Convolutional Network (AGCN)

The skeleton data convolution of the temporal graph described above may not be the optimal approach due to its dependence on an intrinsic human-body network. For this issue, AGCN layers are suggested. Endwise learning enhances graph topology and network features. The graph differs for each layer and sample, making model adjustment easier. Even though it was added as a branch, it keeps the basic model stable. Equation 1 lets them use the GCN in our method.

$$f_{out} = \sum_{k}^{k_v} M_k (f_{in} A_k) \qquad (1)$$

$M_k$ and $A_k$ are Adjacency matrices and masks from the formula. The graph is defined by using Equation 1. $A_k$ checks connections between vertices, and $M_k$ shows the link's strength. Changing Equation 1 way changes the graph's topology. Change Equation 1 to this:

$$f_{out} = \sum_{k}^{k_v} W_k f_{in} (B_k + \alpha C_k) \qquad (2)$$

The graph is identified, and the adjacency matrix is broken into $B_k$ and $C_k$ subgraphs.

The first subgraph ($B_k$) is a data-learned global graph. It represents the best graph topology for action recognition. Equation 1 initializes the human-body-based graph with its adjacency matrix $A_k$. Unlike $A_k$, $B_k$ components are parameterized and changed during training with other parameters. The Bk components in the graph are not limited, indicating that the graph was created using training data. With this data-based method, the model can learn great plots to recognize them. It can handle the different levels of meaning in more than one layer better because $B_k$ is different for each layer.

The second sub-graph ($C_k$) has a unique topology for each sample. We use the normalized embedded Gaussian function to compare the properties of two vertices to find out the link between them and how strong it is. The expression function is written as:

$$f(v_i, v_j) = \frac{e^{\theta}(v)^T \phi(v_j)}{\sum_{j=1}^{N} e^{\theta}(v_i)^T \phi(v_j)} \tag{3}$$

The dot product compares two vertexes where N denotes the vertex number in an implanting setting. Everything is initially implanted into the implanting place $R^{C_e \times N \times T}$, assuming the input feature map $W_2 \in R^{\frac{C}{r} \times C}$ with two embedding functions, denoted by $\theta$ and $\phi$. After extensive experimentation, the embedding function was determined to be the 1 × 1 convolutional layer. The two internal feature maps are changed into two new shapes $M_{\phi K} \in R^{C_e T \times N}$. Then, they are multiplying to get a similar matrix $C_k \in R^{N \times N}$. The component $C_k^{ij}$ shows how similar the vertex $v_i$ and the vertex $v_j$. The matrix's value is set to 0–1, which acts as a soft border between the two places. While the normalized Gaussian does have a SoftMax method, we can use Equation 4 to find out Ck in the following way:

$$C_k = SoftMax(f_{in}^T W_{\theta k}^T W_{\phi k} f_{in}) \tag{4}$$

where $W_\theta \in R^{C_{in} \times C_e \times 1 \times 1}$ represent the corresponding limits of the implanting functions $\theta$ and $\phi$.

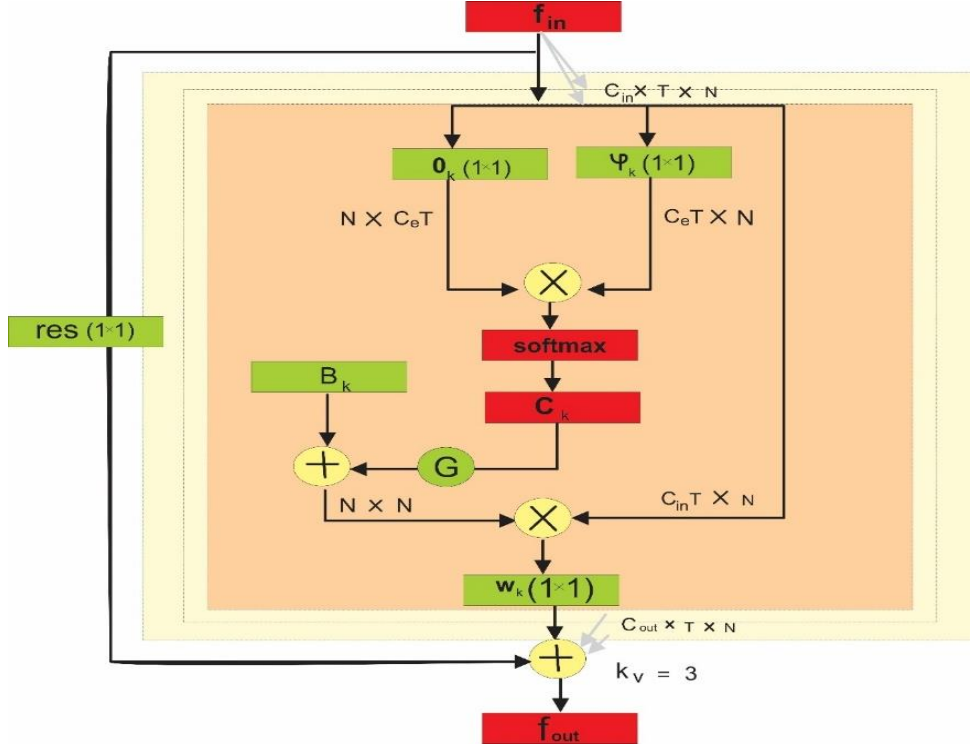

**Figure 1:** An adaptive graph convolutional layer architecture. Each layer has two graph types, $B_k$ and $C_k$. The orange box indicates training adjustments to this parameter. Symbols represent both embedding functions, with a kernel size of *(1x1)*. $K_v$ represents the total subgroups. The ⊕ symbol signifies adding an element, while ⊗ the symbol represents matrix multiplication. The gate determining which graphs are more significant than others is *G*. It's only required in the residue scenario (dashed line) where $C_{in}$ is less than or equal to $C_{out}$.

***Gating mechanism:*** The global graph establishes the fundamental graph topology for action recognition, while the individual graph provides uniqueness based on numerous sample properties. Our studies found that the individual graph is more important in the top and lower layers. The smaller receptive field of the lower layer makes it more challenging to learn the graph structure from samples. The top layers contain more profound and significant information, making the graph topology more changeable and requiring more innovation. Since graphs are created from input properties and are unique to each sample, meeting the condition is more accessible. Utilizing a gating mechanism based on these discoveries changes the graph's importance for different layers. A layer-specific parameterized coefficient multiplies $C_k$. This layer-specific coefficient is learned and changed throughout training.

***Initialization:*** The results of the experiments show that the graph shape changes a lot early on in training. Less stability and convergence issues led to multiple attempts before discovering two working models. Firstly, $A_k+\alpha B_k+C_k$ is used as the adjacency matrix. $A_k$ is the human-body-based graph that stays the same during the process. All $B_k$, $C_k$, and values have been changed to 0. This method will help to regulate the $A_k$ during the initial stages of training. Establish the second approach, $B_k$, in conjunction with $A_k$, and halt the incline's progression for $B_k$ until the initial stages of the training regimen achieve stability.

The entire design of this layer is shown in Figure 2. With $K_v=5$, the graph convolution kernel is ready to go, and the increment task is $W_k$.

A new association, similar to [33, 34], is introduced for each level to allow the layers to be added to existing models without impacting their original performance. Assume there are more input networks than output networks. To fit the output in that circumstance, the remaining route that targets the input will include a *1x1* convolution (indicated by a dotted line in Figure 2). *G* acts as a gatekeeper for two types of graphs.

### 3.3 STC-attention Module

There have been many noticed module compositions [35, 36]. Figure 3, which depicts the STC-attention module, is based on the results of evaluations. All three types of attention—spatial, temporal, and channel —are represented.

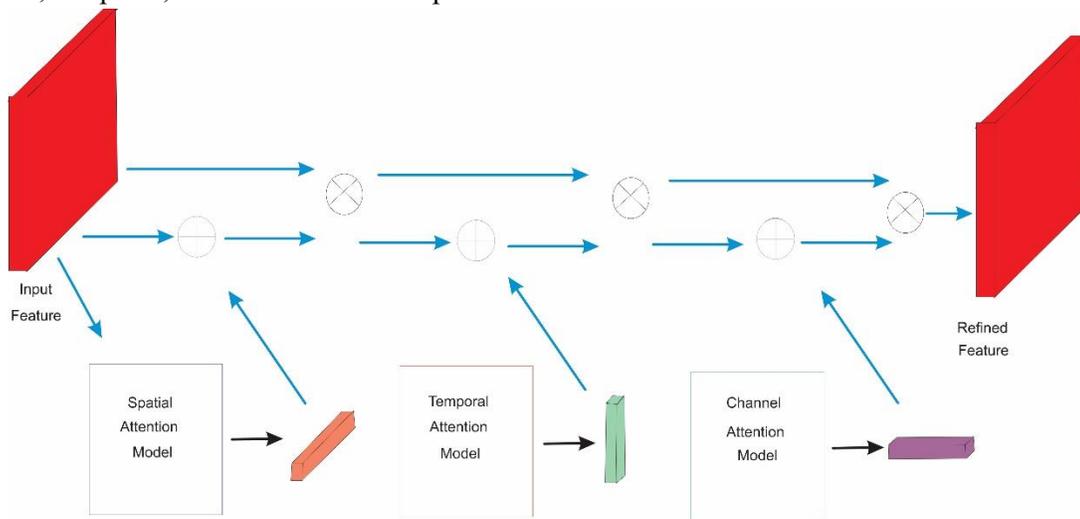

**Figure 3:** STC attention model. Sequencing sub-modules are SAM, TAM, and CAM—multiplication indicated by $\otimes$ while addition indicated by $\oplus$.

**S*patial attention module (SAM):*** The model can receive assistance by paying varied attention to each joint through the SAM. It is computed in the following manner:

$$M_s = \sigma(g_s(AvgPool(f_{in}))) \quad (5)$$

The input feature map is denoted by $f_{in} \in R^{C_{in} \times N \times T}$ and then averaged across all frames. In terms of convolutional operations, the $g_s$ procedure is one-dimensional. σ denotes the sigmoid activation function. $W_{g_s} \in R^{K_s \times C_{in} \times 1}$ where $k_s$ is the kernel size. The attention map $M_s \in R^{N \times 1 \times 1}$ is then multiplied in a residual manner by the input feature map. Adaptive feature modification operates in this manner.

***Temporal attention module (TAM):*** The TAM module was created for effective temporal modeling; Figure 4 illustrates the TAM module's architecture. Consider the structure of the following input feature F: [B, C, T, P, M], where the batch number (B), feature channels (C), temporal dimension (T), and spatial resolutions (P and M) are all defined. While ignoring specific spatial layouts, we use global spatial average pooling to highlight data sensitive to motion.

$$F^S = Pool(F), F^S \in R^{B \times C \times T \times 1 \times 1} \quad (6)$$

Remember that TAM represents two branches. The feature $F^S$ from (6) is then fed to the short-term and long-term branches independently.

***Short-term Branch:*** The objective of the short-term branch is to collect sensitive data on local movements. To reduce the computational load, we implement a *1D* convolutional layer that decreases the channel capacity by r (r = 4).

$$F^r = Conv * F^S, F^r \in R^{B \times \frac{C}{r} \times T \times 1 \times 1} \quad (7)$$

A *1D* convolution kernel with a size of 5 is used., which allows it to learn the temporal information within the immediate district. An additional *1D* convolutional layer, called Convexp, is added to raise the dimension of motion features to *C*. Following this, motion-attentive weights are applied. A Sigmoid function represents by W:

$$W = Sigmoid(Conv * F^r), W \in R^{B \times C \times T \times 1 \times 1} \quad (8)$$

The following procedure is used to access the motion-sensitive data: the input feature and motion-attentive weights are multiplied channel-wise as:

$$F^0 = W \otimes F \quad (9)$$

The output is denoted as $F^0$. A channel-wise multiplication is indicated by $\otimes$. The motion pattern has been thoroughly examined and emphasized using the abovementioned techniques.

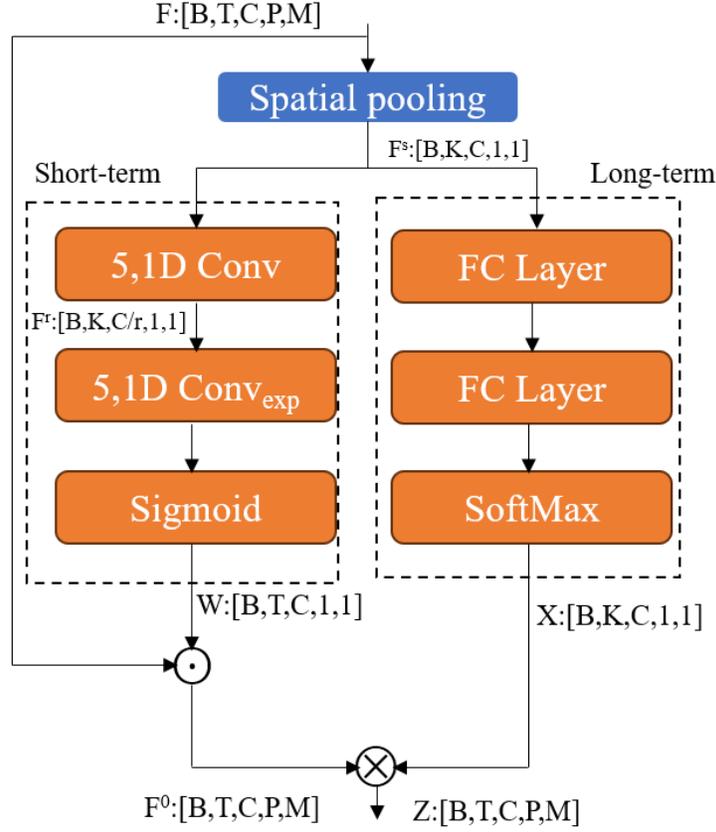

**Figure 4:** The architecture of the TAM module.

***Long-term Branch:*** Unlike the short-term branch, the long-term branch employs a convolution kernel to integrate temporal information but does not calculate channel correlation. As shown in Figure 4, we can learn temporal relations by stacking two fully connected (FC) layers on top of one another. A SoftMax function is used to normalize aggregation weights, defined as follows, and applied after the second fully connected (FC) layer:

$$Z = SoftMax(K(M_2, K(M_1, F_C^S))) \qquad (10)$$

where $Z \in R^{B \times C \times T \times P \times M}$ is the estimated weight for the $c^{th}$ channel, and K = 5 represents the kernel size. The learned weights Z = {$z_1, z_2,...,z_c$} are used. to aggregate the short-term temporal information Formally,

$$Z = X \otimes F^0 \qquad (11)$$

where $\otimes$ is a channel-wise convolutional operator that ensures that information from distinct channels is not mixed. The final feature map is: $Z \in R^{B \times C \times T \times P \times M}$. TAM module acquires both local motion information and global temporal aggregate information using short-term and long-term branches in the preceding steps.

***Channel Attention Module (CAM):*** CAM is a module that can help models improve their discriminative features, or channels, after receiving input samples. As a result, the following attention maps are created:

$$M_c = \sigma(W_2(\delta(W_1(AvgPool(f_{in}))))) \qquad (12)$$

where $M_c \in R^{C \times 1 \times 1}$ is the mean of all joints and frames, and $f_{in}$ is the average of all joints and frames. Two fully connected layers' weights are $W_1 \in R^{C \times \frac{C}{r}}$ $W_2 \in R^{\frac{C}{r} \times C}$, $\delta$ represents the ReLu activation function.

***The attention modules are set up:*** The three submodules mentioned earlier can be arranged sequentially or in parallel, although in various orders. Our findings indicate that the sequential strategy, wherein the SAM, TAM, and CAM procedures are carried out in a specific order, is the most efficient.

## 4. Experimental Setting

We test four comprehensive HAR datasets to compare with ST-GCN directly: NTU [37], Kinetics [38], UCF-101 [14] and HMDB-51 [15]. The NTU dataset is smaller than the Kinetics dataset, so we conducted extensive ablation experiments to validate the model components' recognition performance. The finished model is tested on both datasets for generality and compared to existing top skeleton-based HAR techniques. At last, we integrate skeleton and pose-guided reduced RGB data to increase NTU accuracy.

### 4.1 Network Architecture

The network architecture is shown in Figure 6, which shows how these essential building blocks are arranged (Figure 5). There are nine blocks in all. The following pictures show the output networks for each chunk: 64, 64, 64, 128, 128, 128, 256, 256, 256, 512, 512, and 512. At the start, a Batch Normalization (BN) layer is used to make the receiving data and information more consistent. In the last part, an average global pooling layer combines feature maps from different data into one dimension. Based on the final result, SoftMax classifiers are used to figure out the forecast.

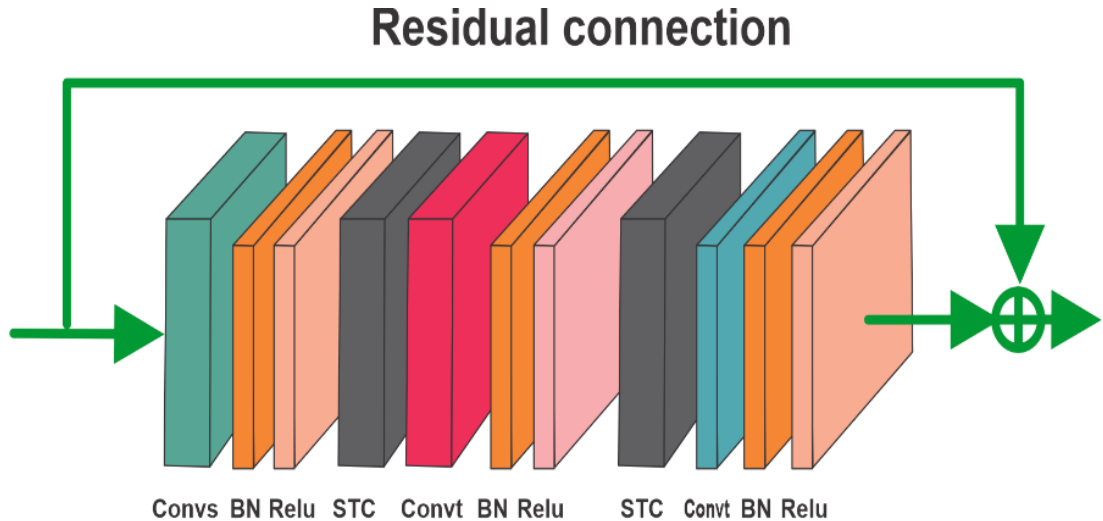

**Figure 5:** This illustrates the basic building block. The notation Convs stands for the spatial AGCL, and the notation Convt stands for the temporal AGCL. A BN layer and a ReLU layer keep track of these notes. STC is the STC-attention module. Each block also has a residual material link.

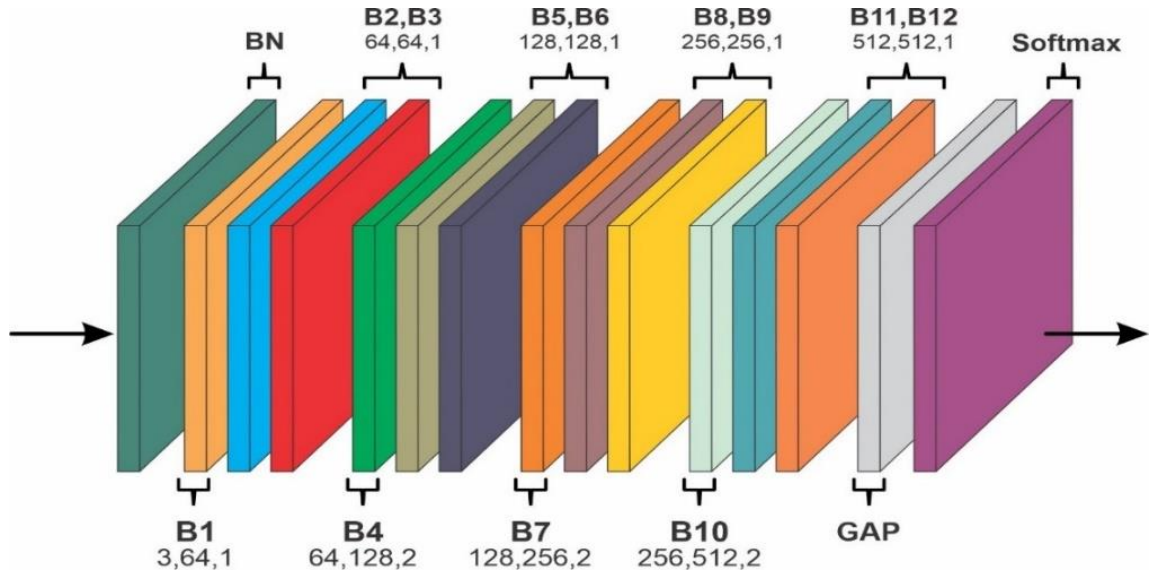

**Figure 6:** Shows network architecture. Nine construction blocks (B1-B9) exist. These three integers represent each block's input, output, and stride. GAP stands for "global average pooling layer."

## 4.2 Training Details

The PyTorch deep learning framework is used for experiments [39, 40]. Stochastic Gradient Descent (SGD) with Nesterov momentum (0.9) was used as an optimization method—batch size is 32. Gradients are a loss function in cross-entropy backpropagation with a weight loss 0.0001.

The NTU dataset contains samples with a maximum of two people. The second body value is 0 if the model has less than two bodies. There can be no more than 300 frames in a sample. We iterate the models for portions with fewer than 300 frames until they reach 300. The current rate is obtained by dividing the learning rate (0.01) for the 30$^{th}$ and 40$^{th}$ epochs by 10. The 50$^{th}$ period ends with the completion of the training.

The input tensor and the kinetics input have the same size. [25], Each included 150 frames with two bodies using the Kinetics dataset. We employ the same methods as in [23] for data augmentation. The initial skeleton sequence is randomly chosen for a duration of 150 frames. Subsequently, the joint coordinates are slightly altered by applying random rotation and translation operations. The learning rate is fixed at 0.001 for the 45$^{th}$ and 55$^{th}$ epochs, subsequently reduced by a factor of ten. The training program terminates after the 65$^{th}$ epoch.

## 4.3 Datasets

*NTU-RGBD:* The NTU [37] dataset, which has 56,000 action clips in 60 action classes, is the biggest and most popular in-door action classification dataset. People between 10 and 35 volunteered to act in the clips. Three cameras at the same height are angled at −45°, 0°, and 45° to record each motion. This data set shows the 3D joint locations of every frame that Kinect depth sensors picked up. Each skeleton sequence comprises 25 joints per subject, with no more than 2 subjects in each video. The original papers [37] suggest two standards: (1) Cross-subject (CS): A training set of 40,320 videos and a validation set containing 16,560 videos make up this benchmark's dataset. The subjects in each part are different. Regarding Cross-view (CV), this benchmark's training set has 37,920 videos taken by 2,3 cameras, while the validation set has 18,960 videos taken by camera 1. We always follow this rule and give the top-1 score in both standards.

***Kinetics-skeleton:*** Kinetics [38] is a massive collection of HAR data that includes 300,000 video clips classified into 400 classes. The various types of video clips were obtained from YouTube videos. The content is made up entirely of raw video samples with no associated skeletal data. S. Yan *et al.,* [25] using the freely accessible OpenPose toolkit, guess the places of 18 joints on each clip [3, 13]. Two people are chosen for multi-person clips using the joint confidence average. We utilize the public data from the Kinetics skeleton to test the proposed network. Both the training and validation sets have 240,000 clips. Adding 240,000 clips to the validation set brings the total to 20,000. We first train the models using the training group to assess them according to the approach described in [25]. We use the training set to train the models, and the validation set reports the top-1 and top-5 accuracy.

***UCF-101:*** There are now more than 13,000 videos in the folder, with 101 distinct action categories consisting of: Each video has a resolution of 320 × 240 pixels and a frame rate of 25 frames per second [14]. Simultaneously, the AlphaPose toolkit was used to capture about 16 joint movements using RGB videos. Similarly, choreographed motions like 'Cutting in the kitchen' closely resemble the specified activities and objects in UCF101. There are a total of 3170 videos showcasing various poses, which are categorized into 24 different classes. These classes include activities such as jump roping, piano playing, crawling baby, flute playing, cello playing, punching, tai chi, boxing speed bag, pushups, juggling balls, golf swing, clean jerk, guitar playing, bowling, ice dancing, playing soccer and juggling, playing dhol and tabla, boxing punching bag, salsa spins and hammer.

***HMDB-51:*** This dataset consists of 6766 video segments that are distributed across 51 action types [15] . Its diversity is widely recognized; it encompasses a wide range of human activity by incorporating acts from numerous sources. The dataset is typically composed of three distinct sets: split 1, split 2, and split 3. Each division ensures that models are assessed on distinct samples by utilizing a distinct configuration of test and training sets.

## 5. Result Analysis

Using the NTU and Kinetics datasets, we compare the resulting model to the most advanced skeleton-based HAR approaches. The results are shown in Tables 1 and 2. For comparisons, Handcraft-feature-based methods [17, 20], RNN-based methods [24, 41, 42], CNN-based methods [34, 43], and GCN-based methods were used. Our model outperforms the existing state-of-the-art on both datasets, indicating that our technique is more successful.

***NTU-RGBD:*** We observe the established CV and CS conventions when evaluating our model on the NTU dataset. Our method is systematically compared with deep learning methodologies that integrate CNNs, RNNs, and LSTM in conjunction with skeleton data, as exemplified by previous works [5, 41, 42]. We contrast our method with traditional methods incorporating graph convolutional elements [25, 44]. The NTU dataset, featuring extensive viewpoints, poses a challenging environment for action recognition. In this context, our proposed models, EMS-TAGCN, surpass conventional configurations and other graph-based attention systems that employ innovative techniques, such as attention mechanisms. Across both CS and CV scenarios, our designed method, EMS-TAGCN, exhibits superior performance, outperforming the best state-of-the-art outcomes and establishing its efficacy in addressing the complexities associated with action recognition in the NTU dataset. Table 1 and Figure 7 demonstrate the results compared to state-of-the-art methods.

Table 1: Accuracy comparisons with other methods using NTU.

| Methods | Years | CS (%) | CV (%) |
|---|---|---|---|
| HBRNN [22] | 2015 | 59.10 | 64.00 |
| ST-LSTM [41] | 2016 | 69.20 | 77.70 |
| TCN [45] | 2017 | 74.30 | 83.10 |
| Clips+CNN+MTLN [5] | 2017 | 79.60 | 84.80 |
| ST-GCN [25] | 2018 | 81.50 | 88.30 |
| Ind-RNN [42] | 2019 | 81.80 | 88.00 |
| GCN-NAS [9] | 2020 | 89.40 | 95.70 |
| TA-GCN [44] | 2021 | 89.91 | 95.8 |
| BRL[46] | 2022 | 91.20 | 86.80 |
| LKA-GCN [47] | 2023 | 90.70 | 96.10 |
| **EMS-TAGCN (our)** | -- | **91.30** | **97.50** |

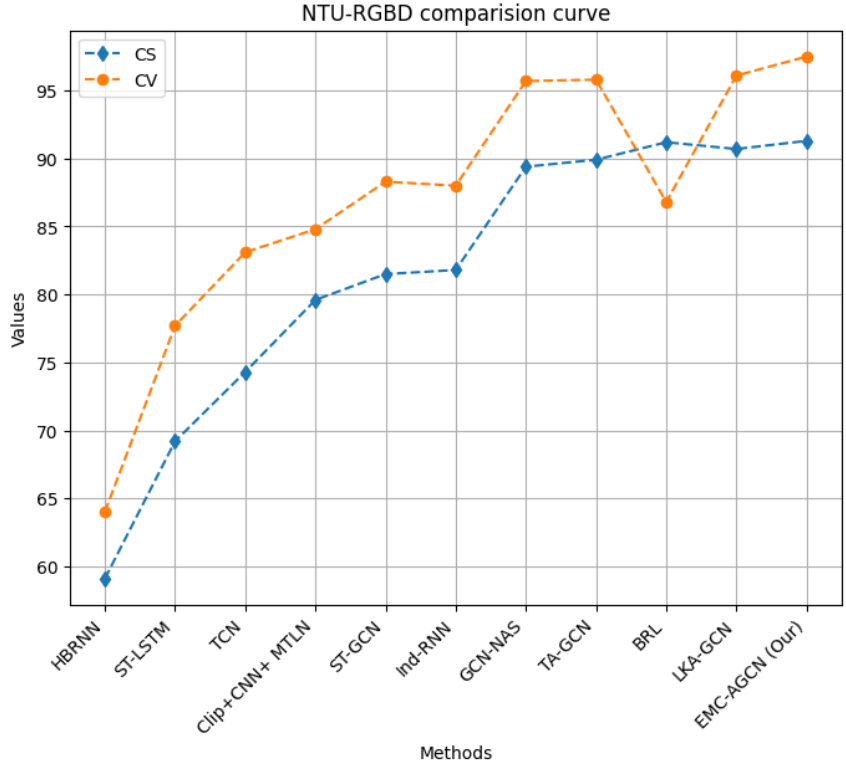

Figure 7: NTU comparisons curve with other methods.

***Kinetics-skeleton:*** We checked how well the EMS-TAGCN method works on the kinetic-skeleton dataset. We looked at the Top-1 and Top-5 results, common standards for evaluating performance. Our process was compared with other deep learning methods that use GCN [47], TCN [45], and LSTM [37], along with skeleton data, as shown in earlier studies. Let me simplify that for you: The Kinetics dataset is quite challenging for recognizing actions because it has many different views. Our proposed models, EMS-TAGCN, are better than usual setups and other attention systems that use special techniques. In both Top-1 and Top-5 situations, EMS-TAGCN works well. It outperforms the best results we had before and proves it's practical for figuring out actions in the

Kinetics dataset, which are shown in Table 2 and in Figure 8, there is a line that compares our proposed model with other models.

Table 2: Accuracy comparisons of other methods using kinetics.

| Methods | Year | Top-1(%) | Top-5(%) |
| --- | --- | --- | --- |
| Feature Enc. [20] | 2015 | 14.90 | 25.80 |
| Deep LSTM [37] | 2016 | 16.40 | 35.30 |
| TCN [45] | 2017 | 20.30 | 40.00 |
| ST-GCN [25] | 2018 | 30.70 | 52.80 |
| DGNN [48] | 2019 | 36.90 | 59.60 |
| GCN-NAS [9] | 2020 | 37.10 | 60.10 |
| TA-GCN [44] | 2021 | 36.10 | 58.720 |
| MS-G3D [49] | 2022 | 38.00 | 60.90 |
| LKA-GCN [47] | 2023 | 37.80 | 60.90 |
| **EMS-TAGCN (our)** | -- | **39.20** | **62.30** |

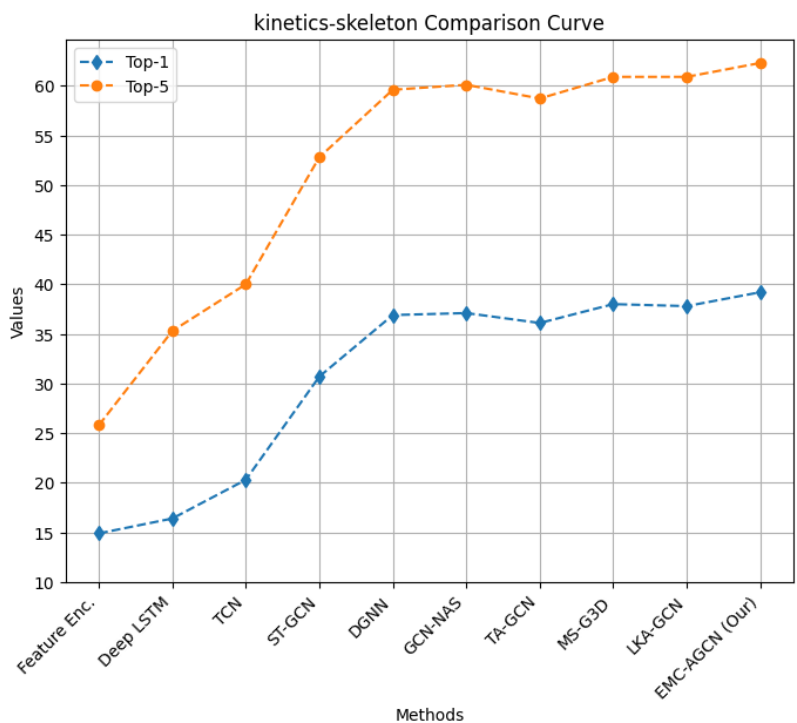

Figure 8: kinetics comparisons curve with state-of-the-art methods.

*UCF-101:* Table 3 indicate that the EMS-TAGCN model outperforms other models examined in this study in successfully completing the UCF-101 dataset. The accuracy of this specific model is the highest, attaining 51.24%, surpassing TSRL (2022) which had an overall accuracy of 48.9%. The ER model from 2021 achieved a maximum accuracy of 47.0%. In comparison, the PS-GNN model from 2020 and the TS-GCN model from 2019 performed worse, with accuracy scores of 43.0% and 34.2% respectively. It is remarkable that the most recent ETSAN (2023) accurately replicates all other criterion facets; however, its overall accuracy level is only 20.6%. This includes the observation that the sophistication of the model does not necessarily equate to higher accuracy.

Based on the aforementioned findings, EMS-TAGCN demonstrates superior performance, which may be ascribed to its capacity to boost action recognition accuracy more effectively than previous techniques.

**Table 3:** Accuracy comparisons of other methods using UCF-101.

| Models | Year | Accuracy % |
|---|---|---|
| TS-GCN [50] | 2019 | 34.2 |
| PS-GNN [51] | 2020 | 43.0 |
| ER [52] | 2021 | 47.0 |
| TSRL [53] | 2022 | 48.9 |
| ETSAN [54] | 2023 | 20.6 |
| **EMS-TAGCN (our)** | --- | **51.24** |

Figure 9 further improves the visual representation of the performance of each model, with the suggested EMS-TAGCN model exhibiting the highest accuracy of 51.24%. The percentage is 24%. This is different from other models such as TS-GCN, PS-GNN, ER, TSRL, ETSAN, and others. The convergence curve and markers show how accuracy has improved across the board, especially for the EMS-TAGCN, which is the most accurate approach.

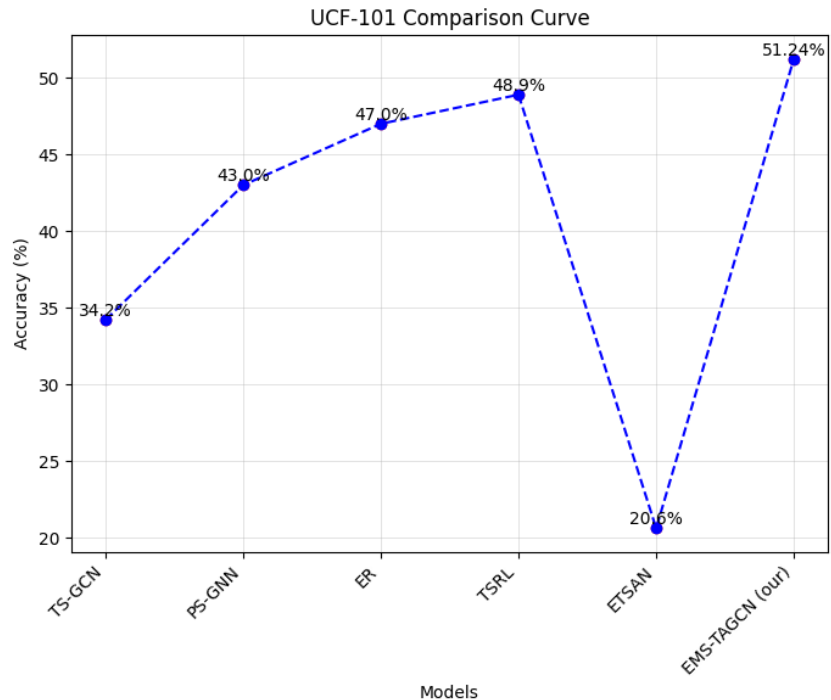

**Figure 9:** UCF-101 comparisons curve with state-of-the-art methods.

*HMDB-51:* Table 4 illustrates the diverse outcomes acquired by different models on HMDB-51 in order to assess the advancements made in action recognition. In terms of accuracy, EMS-TAGCN outperforms the SKP model by over 2% and also surpasses previous studies like PoseC3D and OpenPose + ST-GCN. Therefore, EMS-TAGCN demonstrated exceptional performance in action recognition on the HMDB-51 dataset. Nevertheless, the aforementioned table clearly demonstrates that EMS-TAGCN exhibits a greater level of operational intricacy and effectively offsets this increased burden by achieving unparalleled accuracy, a feat that remains unmatched by any of the current models.

Table 4: Accuracy comparisons of other methods using HMDB-51.

| Models | Year | Accuracy % |
|---|---|---|
| Potion [55] | 2018 | 43.70 |
| GAGCN [56, 57] | 2020 | 32.50 |
| OpenPose + ST-GCN + Index Split [58] | 2021 | 47.69 |
| PoseC3D [59] | 2022 | 56.60 |
| SKP [60] | 2023 | 70.90 |
| **EMS-TAGCN (our)** | --- | **72.70** |

Figure 10 shows the comparison of accuracy levels among the models used with the HMDB-51 dataset. The chart indicates the performance rate of each model, with yours, EMSTAGCN, achieving the highest accuracy of 72.70%, which is better than SKP, PoseC3D, OpenPose and its variations ST-GCN and Index Split, as well as Potion and GAGCN. This is clearly shown by the markers and the convergence curve, which depict the increasing accuracy of different approaches. Notably, EMS-TAGCN achieves the highest score.

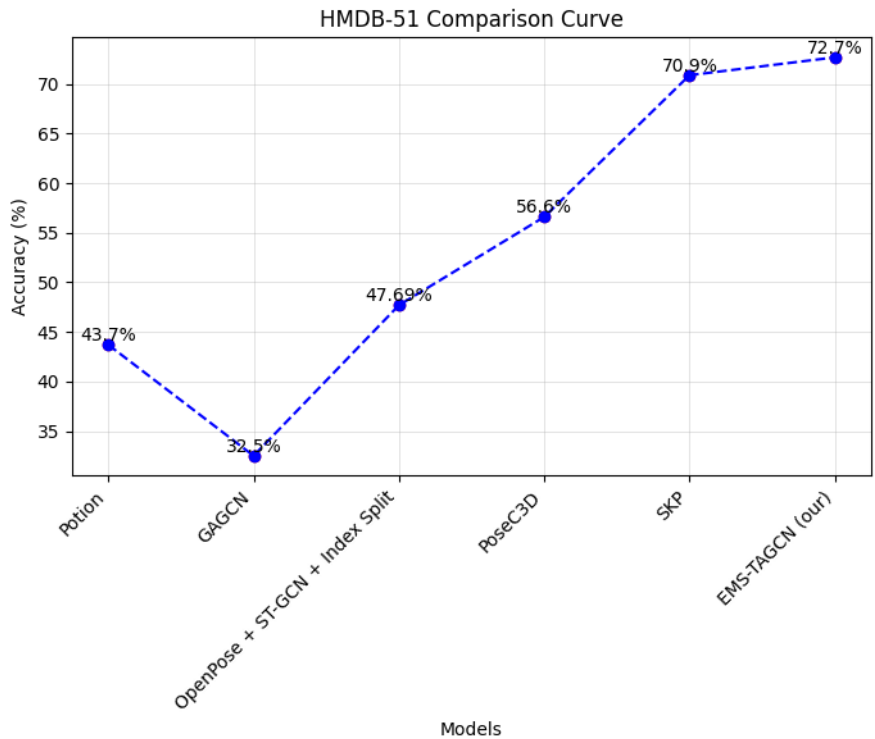

**Figure 10:** HMDB-51 comparisons curve with state-of-the-art methods.

### 5.1 Fusion using the RGB Model

Skeletal data is robust in the face of dynamic conditions and complex environments. It does not, however, include any details about the appearance. For instance, based only on skeletal data, it can be challenging to determine whether someone is eating an apple or a peal when they are witnessed eating anything. This section investigates whether the NTU dataset-based action recognition challenge necessitates the integration of RGB and skeletal data. A two-channel system is utilized, with one channel representing the RGB data using 3D CNN and another employing our EMS-TAGCN to model the skeletal data. The ResNeXt3D-101 model represents the RGB channel. MobileNet and EfficientNet have already been used to train this model. We randomly select a segment of the entire film and end it randomly during training [32,64,128]. The crop placement is chosen randomly from the center and the four sides. Each image is then divided in half using the crop percentage that was previously sampled.

The cropped image's width and height might not match the original dimensions. The sliced photo series is then normalized and scaled to [16,224,224], where [16] is the width, [224] is the height, and [224] is the length of the clip. To achieve an absorption proportion of 0.01, it is first multiplied by 0.1 after the peak of authentication accuracy. We use four TITAN XP-GPUs with a bunch size of 32 to handle the data. With momentum-SGD as the optimizer, we could reduce weight decay to 0.0005%. We will crop clips at various moments and with varying durations throughout the testing session. The outcome is established by averaging these clips.

Additionally, to mitigate the impact of background interference, we suggest cropping individuals from the source photographs and identifying solely the image region that encompasses the individuals. More precisely, the union of the bounds of the individuals in each image is calculated and subsequently employed as the crop box. The pose-guided cropping strategy is suitable for the cropping mentioned above methodology.

As shown in Table 5, the outcomes from our methods are shown along with those from older methods that also used the RGB modality. In this case, the letter C stands for the pose-guided cutting method. When only the RGB mode is used (RNX3D101 instead of RNX3D101-C), the pose-guided cutting method makes a big step forward. Complex backgrounds can fool RGB-based models due to their design limitations. Understanding these weaknesses is crucial to producing accurate results. When the RGB and skeleton data are merged (RNX3D101+EMS-TAGCN instead of RNX3D101+EMS-TAGCN-C), on the other hand, the improvement is lessened. The skeletal data can avoid the scenario's interference, which is why this is the case. As a result, the cropping technique's influence is less visible than it was previously. As shown in Table 5, our top model, RNX3D101+MSAAGCN-C, received a CS standard score of 97.3% and a CV standard score of 99.4%. This technique is substantially more effective than the ones that came before it.

Table 5: RGB modality comparisons using the NTU.

| Methods | pose | RGB | CS (%) | CV (%) |
|---|---|---|---|---|
| DSSCA-SSLM [61] | X | X | 74.9 | - |
| Chained Network [62] | X | X | 80.8 | - |
| RGB + 2D Pose [63] | X | X | 85.5 | - |
| Glimpse Clouds [64] |  | X | 86.6 | 93.2 |
| PEM [65] | X | X | 91.2 | 95.3 |
| I3D+RNN+Attention [66] | X | X | 93.0 | 95.4 |
| EMS-TAGCN | X |  | 91.3 | 97.5 |
| RNX3D101 |  | X | 86.2 | 93.4 |
| RNX3D101-C |  | X | 95.3 | 97.9 |
| RNX3D101+EMS-TAGCN | X | X | 96.3 | 98.7 |
| RNX3D101+EMS-TAGCN-C | X | X | 97.3 | 99.4 |

## 5.2 Attention Module

The STC-attention module comprises three sub-modules: spatial, temporal, and channel attention. Figure 11 shows the spatial attention map for different samples and layers. The circle's size means the joint's significance, with the diameter of the process serving as an indicator. It illustrates that the model focuses on head and hand joints.

Additionally, the focus point is not visible at lower levels. Because lower layers have smaller receptive fields. Good attention mappings are complex.

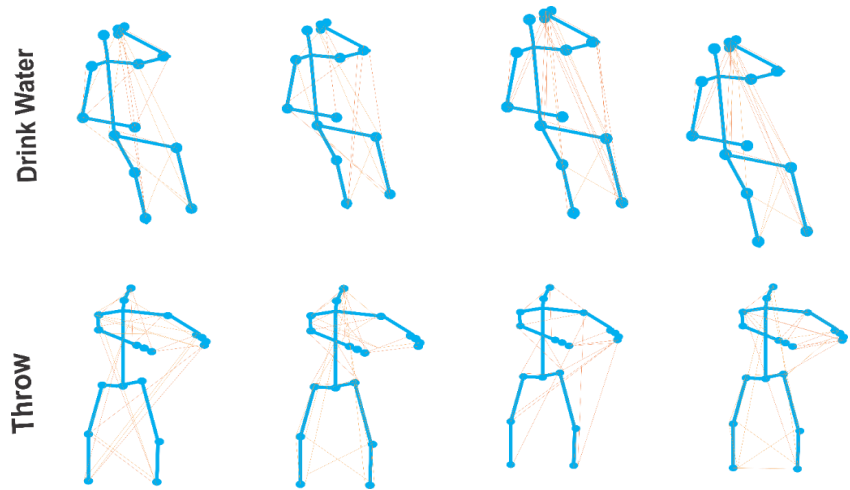

**Figure 11:** Examples of trained spatial attention maps. The circle size denotes joint significance.

In Figure 12, we illustrate the operation of the temporal attention skeleton design linked to the learned attention weights for each frame. The fifth layer of the model pertains primarily to how individuals raise their palms while taking a selfie. Conversely, the seventh layer is more concerned with the final appearance of the selfie. Although it shares the fifth layer with the throwing sample, it investigates a different framework and gives greater attention to the frames where the hands are more buried. This exemplifies the modular design's efficiency and flexibility.

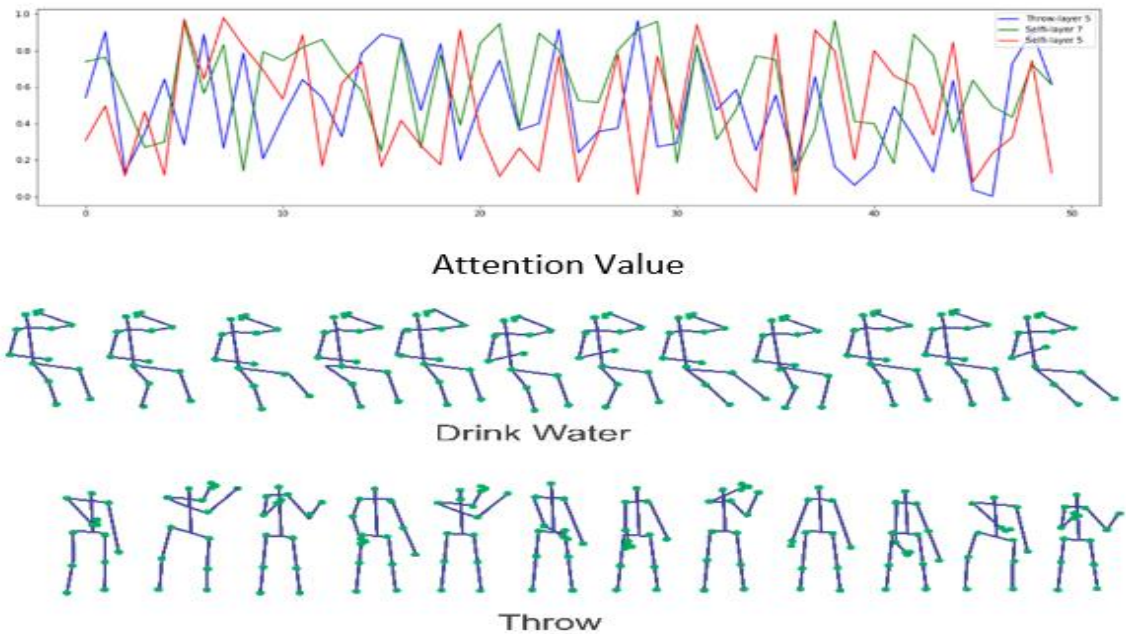

**Figure 12:** The temporal attention map is graphic. The first row for layers and samples shows the temporal devotion masses learned for each frame. Rows two and three contain the second and third-row skeletal sketches.

## 5.3 Multi-modalities

We show the difference in accuracy between the different modalities for some classes to show how different modalities might be able to work together to get more accurate results. Figure 13 demonstrates the accuracy discrepancies that may be observed between the RGB and skeletal modalities. This illustrates the substantial advantages that the skeleton modality imparts to the RGB modality for the "rub the hands" lesson, as well as for the "reading" and "writing" courses. As demonstrated in Figure 14, we come across two distinct occurrences for the classes "reading" and "writing."

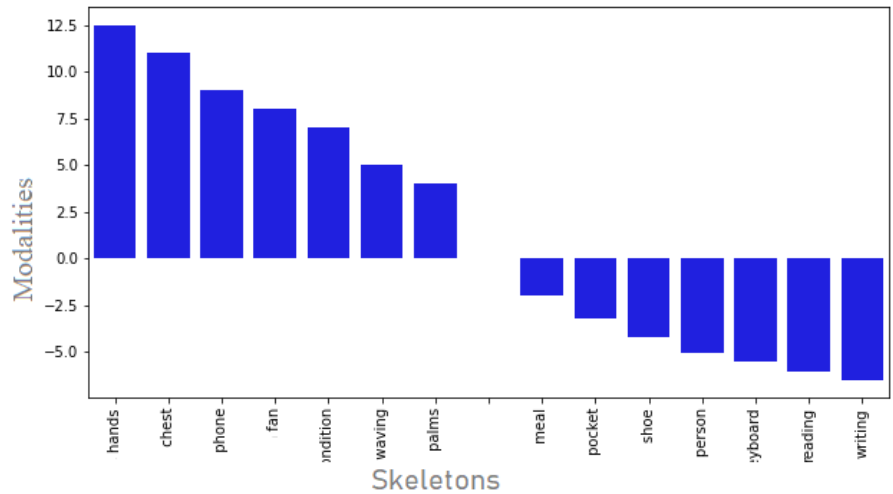

**Figure 13:** Accuracy difference between the skeleton and RGB modalities, i.e., ACC (skeleton)-ACC (RGB).

The skeletons of these two samples are remarkably similar, complicating the distinction. However, using the RGB data makes it possible to distinguish between them based on the presence or absence of a pen in their hands. This graphic explains combining the RGB and skeletal modalities to create a complete image.

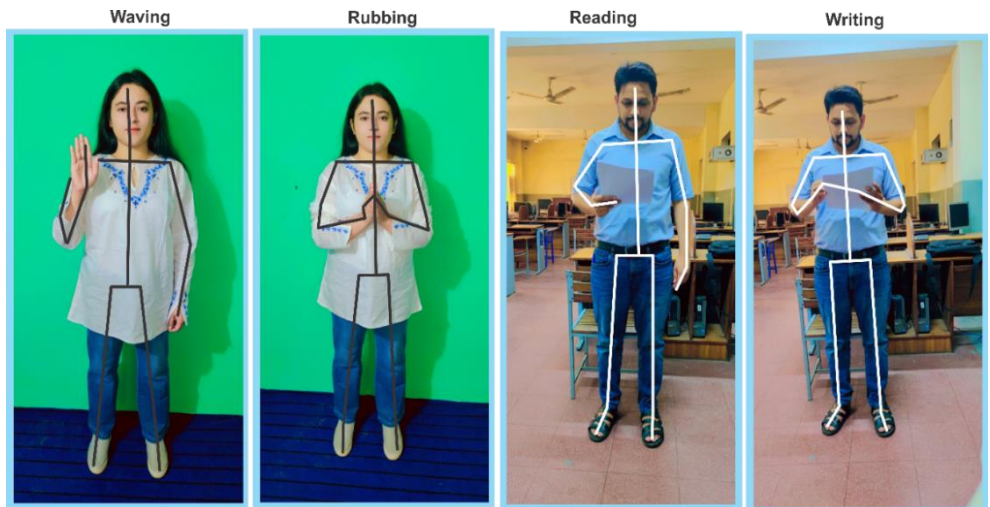

**Figure 14:** Examples of "waving," "rubbing," and "reading," "writing" classes. Skeletal lines are black and white.

# 6. Discussion

This section elaborated the overall significance of the study, results and their implications. In this section we have discussed a summary of study results (section 6.1), the implications of study results (section 6.2) and the future aspect of the study results, see section 6.3.

## 6.1 Results discussion

Our Extended Multi-Stream Temporal-attention Adaptive GCN (EMS-TAGCN) proposed model addresses critical limitations found in previous graph convolutional network (GCN) applications to action recognition. One of the novel contributions of our model is its attention module and adaptive approach to graph topology. Unlike traditional models that rely on a fixed and heuristic graph structure across all layers and datasets, EMS-TAGCN dynamically adjusts its topology. This adaptation is data-driven and occurs end-to-end, greatly enhancing the model's flexibility and applicability to various datasets. This feature is particularly vital given the variability inherent in human action recognition tasks.

Additionally, the incorporation of second-order information—specifically, the lengths and orientations of bones—is a significant advancement. Traditionally, GCN models for action recognition have primarily focused on joint details. By integrating bone sequence, bone motion, bone length: bone (sequence, motion, length) and joint sequence, joint motion, joint length, joint (sequence, motion, length), our model leverages a more comprehensive set of features, which naturally enhances discriminative power and improves recognition accuracy. The use of a spatial-temporal-channel attention module further distinguishes EMS-TAGCN. This module prioritizes crucial joints, frames, and features during the learning process, ensuring that the model focuses on the most informative aspects of the data. The extended multi-stream design of the model is another pivotal aspect. It allows simultaneous modelling of different types of information, such as the sequence, motion and length of joints and bones. This approach not only provides a richer representation of actions but also facilitates the model in capturing indirect variation that might be missed in a single-stream setup.

As a comparative analysis with existing studies, the proposed EMS-TAGCN model perform significantly efficient and the detailed results are given in section 5. For instance, the state-of-the-art BRL model achieved CS and CV scores of 91.20% and 86.80%, respectively, in 2022 while utilizing the NTU dataset. Conversely, the other state-of-the-art LKA-GCN model attained CS and CV scores of 90.70% and 96.10%, respectively, in 2023, also using the NTU dataset. Our EMS-TAGCN model obtained CS and CV scores of 91.30% and 97.50%, respectively, using the same dataset.

Consequently, in 2022, the MS-G3D model achieved Top-1 and Top-5 scores of 38.00% and 60.90%, respectively, employing the kinetics-skeleton dataset. In 2023, the LKA-GCN model achieved Top-1 and Top-5 scores of 37.80% and 60.90%, respectively, using the same dataset. Our EMS-TAGCN model achieved Top-1 and Top-5 scores of 39.20% and 62.30%, respectively. UCF-101 has improved 2.34% accuracy and HMDB-51 dataset has significantly improved 1.8% accuracy.

The effectiveness of EMS-TAGCN is underscored by its performance on prominent datasets like NTU-RGBD, Kinetics-skeleton, UCF-101 and HMDB-51, where it not only meets but significantly surpasses existing benchmarks. This performance boost confirms the value of the adaptive and attentive mechanisms introduced in the model, marking a substantial step forward in the field of action recognition.

Overall, the EMS-TAGCN model introduces several innovative mechanisms that collectively enhance the accuracy and applicability of skeleton-based action recognition. Its adaptive graph structure, integration of second-order information, attention mechanisms, and extended multi-stream framework set new standards for what is achievable with GCN-based models in this domain.

## 6.2 Study Implications

The EMS-TAGCN model represents a significant contribution to the fields of graph neural networks and computer vision, particularly in the application of action recognition. Our work addresses several critical challenges in the existing models and provides a robust framework that could influence future research directions. It also opens up possibilities for real-world applications, such as in surveillance, interactive gaming, sports analysis, and human-computer interaction, where accurate and efficient action recognition is crucial. These points underline the broad and impactful nature of our work, placing it as a pivotal contribution to ongoing research in machine learning and its applications to complex, real-world problems.

This research introduces new and flexible neural network architectures for the exploration of adaptive models. This also indicates potential areas of further study—for example, multimodal data integration or a different type of attention mechanism. Pragmatically, these results are essential for professionals in the field. The presented model proposes robust solutions to actual applications that require exact, accurate, on-time action recognition, such as in the fields of surveillance, robotics, and interactive systems. These explanations are purposed to offer more profound insight into each of the sections of the document, thereby reflecting the comprehensive scope and impact of the research.

## 6.3 Proposed Model Complexity

Out of all the models, four have shown the highest level of complexity: the RNX3D101 + EMS-TAGCN model. It consists of 41.8 million parameters and operates at a pace of 110.5 sequences per second, with 2.11 billion floating-point operations (FLOPs). The integration of the EMS-TAGCN module with the RNX3D101 architecture has resulted in a larger computational complexity and model size compared to the standalone RNX3D101. The EMS-TAGCN module has increased the number of parameters to 17.5 million and the number of FLOPs is 1.7 billion.

Table 6: EMS-TAGCN Model complexity.

| Method | Number of Parameters (M) | Speed (seq/sec.) | FLOPs |
|---|---|---|---|
| EMS-TAGCN | 24.3 | 95.4 | 1.9 billion |
| RNX3D101 | 17.5 | 93.2 | 1.7 billion |
| RNX3D101-C | 15.7 | 92.5 | 1.8 billion |
| RNX3D101+EMS-TAGCN | 41.8 | 110.5 | 2.11 billion |
| RNX3D101+EMS-TAGCN-C | 40.0 | 109.7 | 2.10 billion |

## 6.4 Future Direction

In future, the model could be optimized for practical use in real-time applications like surveillance, interactive gaming, etc. Further tuning the adaptive mechanisms of the model will enable it to have improved performance without so much manual tuning, and this increase in cross-dataset generalization will allow it to have broader applicability across the globe. This technology will find applications in health monitoring and rehabilitation for determining the effectiveness of the offered therapy and diagnosing various movement disorders. Addressing ethics and privacy will be crucial to ensure that technology still flourishes by developing action recognition technologies that are consensual and ethical. Additionally, other relevant future works include extending the effectiveness of this model into different environments and also promoting interdisciplinary collaborations in an attempt to uncover further insights or possible applications in the area of human action recognition.

# 7. Conclusion

This research introduces an extended multi-stream temporal-attention adaptive GCN (EMS-TAGCN) for skeleton-based HAR. In this model, the graph structure of the skeleton data is parameterized and built into the network so that it can be learned and changed along with the other parameters. A data-based approach makes the model flexible and relevant to more circumstances. Also, the human body-based graph beats the adaptively learned topological graph in action recognition. In the convolutional graph, each layer has an STC-attention module that helps the model pay attention to essential joints, frames, and features. We also show motion, joints, and bones in a single extended multi-stream design to make things go faster. Two large action-recognition datasets, NTU-RGB+D, kinetics-skeleton, UCF-101 and HMDB-51 are used to test the final network. We have achieved state-of-the-art performance for both skeletons and skeleton-guided cropped RGB data. Additionally, we have combined the skeleton data with the skeleton-guided cropped RGB data, leading to further improvements. It also enhances the image when skeleton and skeleton-guided RGB data are combined.